# Toward safe separation distance monitoring from RGB-D sensors in human-robot interaction

Petr Švarný, Zdenek Straka, and Matej Hoffmann

*Abstract*— The interaction of humans and robots in less constrained environments gains a lot of attention lately and safety of such interaction is of utmost importance. Two ways of risk assessment are prescribed by recent safety standards: (i) power and force limiting and (ii) speed and separation monitoring. Unlike typical solutions in industry that are restricted to mere safety zone monitoring, we present a framework that realizes separation distance monitoring between a robot and a human operator in a detailed, yet versatile, transparent, and tunable fashion. The separation distance is assessed pair-wise for all keypoints on the robot and the human body and as such can be selectively modified to account for specific conditions. The operation of this framework is illustrated on a Nao humanoid robot interacting with a human partner perceived by a RealSense RGB-D sensor and employing the OpenPose human skeleton estimation algorithm.

## I. INTRODUCTION

As robots are leaving safety fences and begin to share their workspace with humans, they need to dynamically adapt to interactions with people and guarantee safety at every moment. There has been a rapid development in this regard in the last decade with the introduction of new safety standards [1], [2] and a fast growing market of so-called "collaborative robots". Haddadin and Croft [3] provide a recent survey of all the aspects of physical Human-Robot Interaction (pHRI). There are two ways of satisfying the safety requirements for pHRI: (i) *Power and Force Limiting* and (ii) *Speed and Separation Monitoring (SSM)* [2]. In the former case, physical contacts with a moving robot are allowed but need to be within human body part specific limits on force, pressure, and energy. This is addressed by interaction control methods for this *post-impact* phase (see the survey [4]). Safe collaborative operation according to SSM demands that a *protective separation distance*, $S_p$, is maintained between the operator and robot at all times. When the distance decreases below $S_p$, the robot stops [2]. In industry, $S_p$ is typically safeguarded using light curtains or safety-rated scanners.

In this work, we present a framework that combines state of the art solutions and realizes separation monitoring between a robot and a human operator in a detailed, yet versatile, transparent, and tunable fashion. The separation distance is assessed pair-wise for all keypoints on the robot and the human body and as such can be selectively modified to account for various interaction scenarios. The operation

The authors are with the Department of Cybernetics, Faculty of Electrical Engineering, Czech Technical University in Prague (e-mail: petr.svarny@fel.cvut.cz; zdenek.straka@fel.cvut.cz; matej.hoffmann@fel.cvut.cz).

of this framework is illustrated on a Nao humanoid robot interacting in real-time with a human partner who is perceived by a RGB-D sensor.

## II. RELATED WORK

A functional solution for safe pHRI according to SSM will necessarily involve: (i) sensing of the human operator's as well as robot's positions and speeds, (ii) a suitable representation of the corresponding separation distances and (iii) appropriate responses of the machine.

Tracking the spatial location of the robot's keypoints is relatively easy thanks to forward kinematics and joint encoder values. The perception of human operator's location is more difficult. Zone scanners used in industry report the intrusion of an object into a predefined zone—a solution that is safe but very inflexible and essentially prevents most collaborative activities. Two key technologies have appeared recently that facilitate progress in this area: (i) compact and affordable RGB-D sensors (like Kinect) and (ii) convolutional neural networks for human keypoint extraction from camera images [5], [6]. These technologies together—albeit currently not safety-rated—make it possible to perceive the positions of individual body parts of any operator in the collaborative workspace in real time.

Once the robot and human positions are obtained, their relative distances need to be evaluated (see Flacco et al. [7] for a comparison of approaches). The robot and human body parts can be represented as spheres [8], capsules [9] or meshes [10] and they can be different for the robot and the human [11].

The approach is often "robot-centered" in the sense that the collision primitives are centered on the robot body and possibly dynamically shaped based on the current robot velocity [12], [13]. Even the biologically inspired approach to "peripersonal space" representation [10], [11], [14], [15] is robot-centered: the safety margin is generated by a distributed array of receptive fields surrounding the electronic skin of the iCub humanoid robot. Finally, there is a large body of work dealing with motion planning and control in dynamic environments. Most recent and most related to our approach are [9], [11], [16].

We propose a separation distance representation that treats robot and human keypoints equally and uses Euclidean distance in Cartesian space to evaluate all safety thresholds. In accordance with [2], velocities, reaction times, and uncertainties can all flow into the desired thresholds. Unique to our approach, the representation is maximally transparent with the easy incorporation of important features. In opposition to machine learning heavy approaches, our framework allows simple risk assessment and it is straightforwardly transferable between robotic platforms.

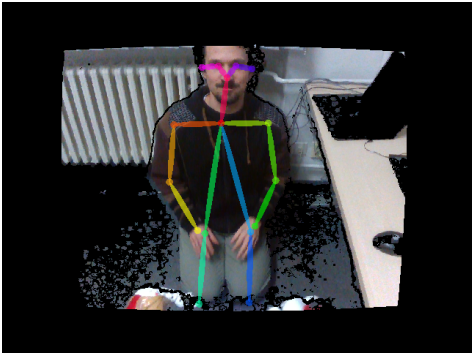

Fig. 1: Color aligned with depth stream with the rendered human keypoints from OpenPose.

## III. MATERIALS AND METHODS

ft Human keypoints are perceived in the environment while robot keypoints are extracted from the model and current joint values. The relative distances are assessed and fed into the robot controller to generate appropriate responses.

### A. Human keypoint 3D estimation

A server collects two streams from a RealSense SR300 camera: a color image aligned to the depth image (CAD) and a point cloud stream (PCS), also depth image aligned. We use Intel RealSense SDK with PyRealSense. The CAD image is sent to OpenPose [5] by PyOpenPose to estimate the human keypoints (see Fig. 1). The pixel coordinates of keypoints are paired with those from PCS. All our image operations use OpenCV3 [17].

The keypoints are transformed into the Nao's frame of reference by affine transforms. The rotation and translation for them are gained from a pre-experiment calibration.

### B. Nao robot keypoints

A Nao humanoid robot (V3+) with keypoints on the left end-effector, forearm, and elbow was used to demonstrate the framework. We used forward kinematics with current joint encoder values as input to get the 3D position of these keypoints.

### C. Separation distance representation

The *protective separation distance* $S_p$ [2] needs to be maintained between any human and robot part such that the human will never collide with a moving machine. Its value will be determined based on reaction times etc. as in [2]. We extend $S_p$ as a baseline with additional terms.

First, we want to account for "modulation" on the part of the human to grant larger distance from specific body parts (e.g. head) and on the part of the robot when carrying a sharp tool. Adding these distance offsets $\mathbf{r}_s$, $\mathbf{h}_s$ gives rise to a *guaranteed minimal separation distance* $S_g$.

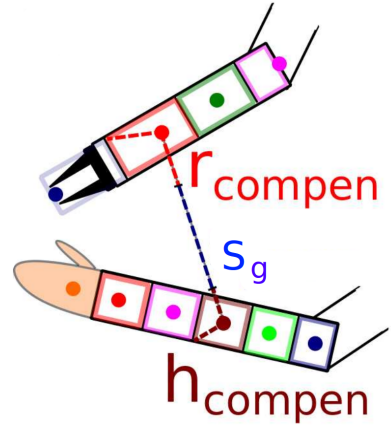

Fig. 2: Separation distance calculation between robot and human keypoints.

Second, as only distances between keypoints will be evaluated, but separation distance between any body parts needs to be maintained, we add compensation coefficients, $\mathbf{h}_{compen}$ and $\mathbf{r}_{compen}$ (see Section III-D below). This is the *keypoint separation distance* $S_d$—the quantity that will be monitored between any keypoint pairs.

Therefore $S_d$ is in the form of a matrix of separation distances between two given keypoints $i$, $j$ ($S_d^{i,j}$) (see Section IV).

$$S_g^{ij} = h_s^i + S_p + r_s^j$$
$$S_d^{ij} = h_{compen}^i + S_g^{ij} + r_{compen}^j$$

### D. Keypoint compensation coefficients

Using a discrete distribution of keypoints allows fast calculation, but does not take the full volume of the bodies into account. The compensation coefficients $r_{compen}$ and $h_{compen}$ allow us to guarantee $S_g$ even with a discrete keypoint distribution.

These coefficients are calculated in two steps. First, every part of the body is assigned to its nearest keypoint. Then the maximal distance over all of its assigned volume is selected as the compensation coefficient for the keypoint (see Fig. 2)—thereby always guaranteeing $S_g$.

### E. Robot control

We used PyNaoqi to control the Nao. The Nao was moving his hands back and forth periodically in front of his chest. The robot stopped when an $S_d^{i,j}$ threshold was exceeded. The robot resumed operation upon "obstruction" removal. In addition, we defined a reduced speed distance: when $S_{d(reduced)}^{i,j}$ for any keypoint pair was exceeded, the robot reduced its speed to half.

### F. HRI setup

The Nao robot was sitting in a fixed position with respect to the camera that captured the robot's workspace (see Fig. 1). Our setup is safe because of the Nao robot's size and power. In a real setting with a potentially dangerous machine and safety-rated modes, $S_p$ would be determined from [2]. In our case, the threshold was chosen arbitrarily.

The compensation values accounting for keypoint density (Section III-D) were determined by measuring the distances between keypoints (Table ?? and I). Only upper body keypoints were taken into consideration for the human operator. We call the set of keypoints of the nose, neck, eyes, and ears as the human head. In both, human and robot cases, the compensation coefficients were symmetrical and thus we list keypoint pairs only once.

| End effector | Wrist | Elbow |
|---|---|---|
| 0.06m | 0.05m | 0.06m |

| Nose | Neck | Eye | Ear | Shoulder |
|---|---|---|---|---|
| 0.10m | 0.25m | 0.10m | 0.10m | 0.15m |
| **Elbow** | **Wrist** | **Hip** | **Knee** | **Ankle** |
| 0.15m | 0.15m | 0.00m | 0.00m | 0.00m |

TABLE I: Human compensation values $\mathbf{h}_{compen}$

## IV. RESULTS

We conducted three scenarios: (A) basic separation matrix, (B) specific separation values for the head of the human, (C) emulation of a sharp tool in the robot's hand.[1] Distances between all human and robot keypoints were evaluated simultaneously online. However, for clarity, we present only the interaction of the robot end-effector with two human keypoints (the right wrist and the nose) in the plots below. The baseline protective separation distance was set to $S_p = 0.05m$ and the reduced speed regime $S_{p(reduced)} = 0.20m$.

### A. Basic scenario

In the basic experiment, we monitored the distance between the human wrist and robot end-effector – see Fig. 3. The relevant separation matrices are in the Table II.

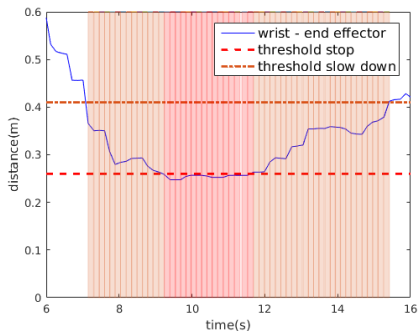

Fig. 3: Basic Scenario: presented are Nao end-effector and human wrist keypoint distances and thresholds ($S_d$ and $S_{d(reduced)}$).

Crossing the threshold into the warning regime is detected by the robot around $t = 7s$ as shown by the orange shaded area. The robot enters reduced speed mode at this point.

[1]The video is available at https://youtu.be/3DZyuuQlqPo.

| | $S_{d(reduced)}$ | |
|---|---|---|
| Robot \ Human | Nose | Wrist |
| **End effector** | 0.36m | 0.41m |

| | $S_d$ | |
|---|---|---|
| Robot \ Human | Nose | Wrist |
| **End effector** | 0.21m | 0.26m |

TABLE II: Basic scenario: Separation matrix for keypoint pairs from Fig. 3.

Similarly, the next crossing is marked by red shading and the robot stops. The removal of the wrist from the safety zones resumes the robot's operations.

### B. Head and body discrimination

The $\mathbf{h}_s$ for the head keypoints was enlarged by $0.15m$. This lead to the robot's higher sensitivity to situations when the human operator approached the robot with his head, as shown in Fig. 4.

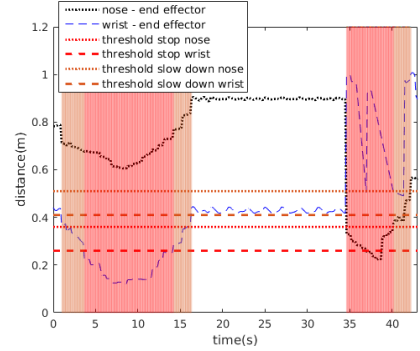

Fig. 4: Head and body discrimination: A higher separation threshold for the human head region.

| | $S_{d(reduced)}$ | |
|---|---|---|
| Robot \ Human | Nose | Wrist |
| **End effector** | *0.51*m | 0.41m |

| | $S_d$ | |
|---|---|---|
| Robot \ Human | Nose | Wrist |
| **End effector** | *0.36*m | 0.26m |

TABLE III: Head and body discrimination: Separation matrix for keypoint pairs from Fig. 4. Emphasis is on values altered w.r.t. to first scenario.

In the first half of the experiment, we see the reaction of the robot to the wrist keypoint. Later, we see that the robot reacts to the nose keypoint at a greater distance than to the wrist. Notice the different reactions of the robot (shown by the different shading) for similar distances of the two keypoints.

### C. Dangerous tool usage

The left arm end-effector $\mathbf{r}_s$ was increased by $0.1m$ to simulate a possibly dangerous tool (see Fig. 5). The stopping and warning thresholds are now $0.1m$ farther away from the robot end-effector. This increase is added to the original functionality from the previous scenario, thus the robot reacts with greater sensitivity to the approach of the operator's nose keypoint as opposed to the proximity of the operator's wrist keypoint.

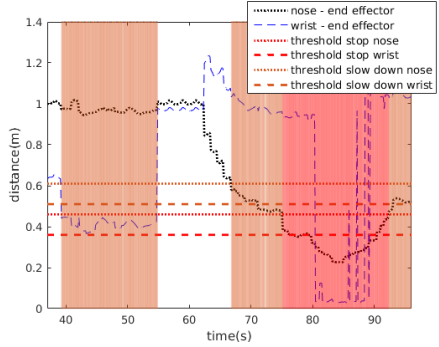

Fig. 5: Dangerous tool usage: Increased safety margin around robot end-effector.

| | $S_{d(reduced)}$ | |
|---|---|---|
| Robot \ Human | **Nose** | **Wrist** |
| **End effector** | **0.61**m | **0.51**m |

| | $S_d$ | |
|---|---|---|
| Robot \ Human | **Nose** | **Wrist** |
| **End effector** | **0.46**m | **0.36**m |

TABLE IV: Dangerous tool usage: Separation matrix for keypoint pairs from Fig. 5.

## V. DISCUSSION AND CONCLUSION

We presented a framework that realizes separation monitoring between a robot and a human operator. Distances are simply represented in Cartesian space in Euclidean norm and human and robot keypoints are treated equally. The separation distance is assessed pair-wise for all keypoints on the robot and human body and as such can be selectively modified. Velocity is not part of our representation but velocities can be converted into distance increments relying on measured quantities or worst-case constants per [2]. The framework was illustrated on a Nao humanoid robot interacting with an operator monitored by an RGB-D sensor.

RGB-D sensors are currently not safety-rated. However, their reliability can be improved [18], [19]. OpenPose itself also provides confidence values with every keypoint estimated. These enhancements and the transfer to a real-life industrial scenario with performance evaluation constitute our future work.

Nevertheless, safety-rated devices similar to those for zone monitoring that would provide 3D object coordinates and possibly human keypoints are needed. Other alternatives exist [20] next to RGB-D sensors. The availability of such technology would expand the possibilities of human-robot collaboration in the SSM regime.

## ACKNOWLEDGMENT

Matej Hoffmann was supported by the Czech Science Foundation under Project GA17-15697Y. Petr Švarný and Zdenek Straka were supported by the Czech Technical University in Prague, grant No. SGS18/138/OHK3/2T/13.